\documentclass[sigconf]{acmart}

\usepackage{subfigure}
\usepackage{multirow}
\usepackage{threeparttable}
\usepackage{graphicx}
\usepackage{textcomp}
\usepackage{xcolor}
\usepackage{enumitem}
\usepackage{comment}
\usepackage{pifont}
\usepackage{amsthm}
\usepackage{algorithmic}
\usepackage{algorithm}
\usepackage{hyperref} 
\usepackage{amsmath}
\usepackage{bm}
\usepackage{bbding}

\DeclareUnicodeCharacter{2217}{}

\AtBeginDocument{%
  }

\author{Yuan~Yuan}
\affiliation{
\institution{Department of Electronic Engineering \\ Tsinghua University \\ Beijing, China}
\country{}
}

\author{Jingtao~Ding$^{*}$}
\thanks{$^{*}$Jingtao Ding is the corresponding author (dingjt15@tsinghua.org.cn, y-yuan20@mails.tsinghua.edu.cn).}
\affiliation{
\institution{Department of Electronic Engineering \\ Tsinghua University \\ Beijing, China}
\country{}
}

\author{Chenyang~Shao}
\affiliation{
\institution{Department of Electronic Engineering \\ Tsinghua University \\ Beijing, China}
\country{}
}

\author{Depeng~Jin}
\affiliation{
\institution{Department of Electronic Engineering \\ Tsinghua University \\ Beijing, China}
\country{}
}

\author{Yong~Li}
\affiliation{
\institution{Department of Electronic Engineering \\ Tsinghua University \\ Beijing, China}
\country{}
}

\renewcommand{\shortauthors}{Yuan Yuan, Jingtao Ding∗, Chenyang Shao, Depeng Jin, \& Yong Li}

\copyrightyear{2023}
\acmYear{2023}
\setcopyright{rightsretained}
\acmConference[KDD '23]{Proceedings of the 29th ACM SIGKDD Conference on Knowledge Discovery and Data Mining}{August 6--10, 2023}{Long Beach, CA, USA}
\acmBooktitle{Proceedings of the 29th ACM SIGKDD Conference on Knowledge Discovery and Data Mining (KDD '23), August 6--10, 2023, Long Beach, CA, USA}
\acmDOI{10.1145/3580305.3599511}
\acmISBN{979-8-4007-0103-0/23/08}

\makeatletter
\gdef\@copyrightpermission{
  \begin{minipage}{0.3\columnwidth}
   \href{https://creativecommons.org/licenses/by/4.0/}{\includegraphics[width=0.90\textwidth]{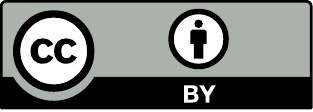}}
  \end{minipage}\hfill
  \begin{minipage}{0.7\columnwidth}
   \href{https://creativecommons.org/licenses/by/4.0/}{This work is licensed under a Creative Commons Attribution International 4.0 License.}
  \end{minipage}
  \vspace{5pt}
}
\makeatother
\begin{document}

\title{Spatio-temporal Diffusion Point Processes}



\begin{abstract}
  Spatio-temporal point process (STPP) is a stochastic collection of events accompanied with time and space. 
  Due to computational complexities, existing solutions for STPPs compromise with conditional independence between time and space, which consider the temporal and spatial distributions separately. 
The failure to model the joint distribution leads to limited capacities in characterizing the spatio-temporal entangled interactions given past events. 
  In this work, we propose a novel parameterization framework for STPPs, which leverages diffusion models to learn complex  spatio-temporal  joint distributions. 
 We decompose the learning of the target joint distribution into multiple steps, where each step can be faithfully described by a Gaussian distribution.  
 To enhance the learning of each step, an elaborated spatio-temporal co-attention module is proposed to capture the interdependence between the event time and space adaptively.
  For the first time, we break the restrictions on spatio-temporal dependencies in existing solutions, 
 and enable a flexible and accurate modeling paradigm for STPPs. 
  Extensive experiments from a wide range of fields, such as epidemiology, seismology,  crime, and urban mobility,  demonstrate that our framework outperforms the state-of-the-art baselines  remarkably. 
  Further in-depth analyses validate its ability to capture spatio-temporal  interactions, which can learn adaptively for different scenarios. 
  The datasets and source code are available online: \textcolor{blue}{\url{https://github.com/tsinghua-fib-lab/Spatio-temporal-Diffusion-Point-Processes}}.
\end{abstract}



\maketitle

\section{Introduction}

Spatio-temporal point process (STPP) is a stochastic collection of points, where each point denotes an event $x=(t,s)$ associated with time $t$ and location $s$. 
STPP is a principled framework for modeling sequences consisting of spatio-temporal events, and has been applied  in a wide range of fields, such as earthquakes and aftershocks~\cite{ogata1988statistical, chen2020neural}, disease spread~\cite{meyer2012space,park2022non}, urban mobility~\cite{wang2021spatio,yuan2023learning,yuan2022activity, long2023VAE}, and emergencies~\cite{xu2016crowdsourcing,zhu2021spatio}.

\begin{figure}[t]
    \centering
    \includegraphics[width=0.95\linewidth]{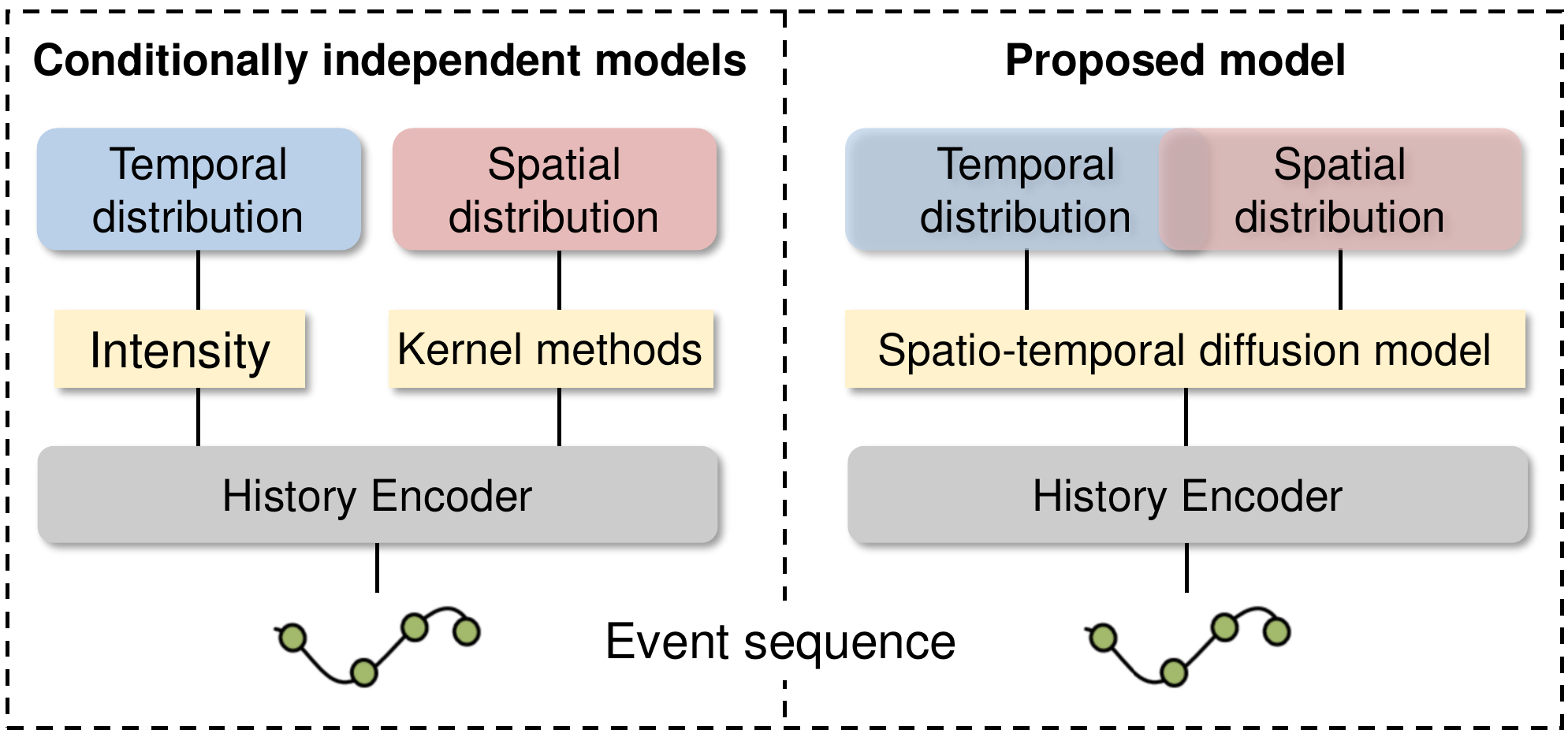}
    \caption{High-level comparison between our proposed framework and conditionally independent solutions for modeling STPPs. Our framework can directly learn the  spatio-temporal joint distribution without any model restrictions.}
    \label{fig:method_cmp}
\end{figure}

Spatio-temporal point processes have been widely studied in the literature~\cite{diggle2006spatio, gonzalez2016spatio, reinhart2018review, baddeley2007spatial, chen2020neural, zhuang2006second, yang2018recurrent} with rich theoretical foundations~\cite{daley2003introduction,grandell2012aspects,kingman1992poisson}. 
Due to computational complexities, a general approach for STPPs is to characterize the event time and space with distinct models. Conventional STPP models~\cite{diggle2006spatio,gonzalez2016spatio, reinhart2018review} mainly capture relatively simple patterns of spatio-temporal dynamics, where the temporal domain is modeled by temporal point process models, such as Poisson process~\cite{kingman1992poisson}, Hawkes process~\cite{hawkes1971point}, and Self-correcting process~\cite{isham1979self}, and the spatial domain is usually fitted by kernel density estimators (KDE)~\cite{wkeglarczyk2018kernel}.
With the advance of neural networks, a series of neural architectures are proposed to improve the fitting accuracy~\cite{jia2019neural,chen2020neural,zhou2022neural}. However, they still adopt the approach of separate modeling. 
For example, Chen et al.~\cite{chen2020neural}  use neural ODEs and continuous-time normalizing flows (CNFs) to learn the temporal distribution and spatial distribution, respectively.  Zhou et al.~\cite{zhou2022neural} apply two independent kernel functions for time and space, whose parameters are obtained from neural networks,  to build the density function.

However, for STPPs, the time and space where an event occurs are highly dependent and entangled with each other. 
For example, in seismology, earthquakes are spatio-temporal correlated due to crust movements~\cite{wang2017earthquake}, which occur with a higher probability close in time and space to previous earthquakes. 
Take urban mobility as another example, people are more likely to go to work during the day, while tend to go for entertainment at night. 
Therefore, it is crucial to learn models that can address the spatio-temporal joint distribution conditioned on the event history.  
However, it is non-trivial due to the following two challenges:

\begin{enumerate}[leftmargin=*]
    \item \textbf{Spatio-temporal joint distributions for STPPs usually have tremendous sample spaces, which are highly intractable.}
    Directly fitting requires huge training samples, which is prohibitive in practice.  
    The general approach is to decompose the target distribution into conditionally dependent distributions~\cite{daley2003introduction,chen2020neural}, fitting the temporal density $p^*(t)\footnote{We use the common star superscript to denote conditional dependence on the history.}$ and conditional density $p^*(s|t)$ separately. 
    However, the characterization of $p^*(s|t)$ is largely limited to certain model structures, such as KDEs and CNFs, which are less expressive. 
    \item \textbf{The occurrence of events is usually associated with complex coupling correlations between time and space.} Driven by different generation mechanisms,  
    the occurrence of events exhibits distinct  spatio-temporal dependencies across various fields.
    How to effectively capture the underlying dependence for an event still remains an open problem. 
\end{enumerate}

Solving the above two challenges calls for a new modeling paradigm for STPPs. 
In this paper, we propose a novel parameterization framework,  \underline{S}patio-\underline{T}emporal \underline{D}iffusion \underline{P}oint \underline{P}rocesses (DSTPP), which is capable of leaning spatio-temporal  joint  distributions effectively. 
By leveraging denoising diffusion probabilistic modeling, we manage to decompose the original complex distribution into a Markov chain of multiple steps, where each step corresponds to a minor distribution change and can be modeled faithfully by a Gaussian distribution~\cite{sohl2015deep,rackwitz1978structural}. 
The target distribution is learned throughout the combination of all steps, where the predicted joint distribution obtained from the previous step acts as the condition for the next-step learning.  
In this way, conditioned on the already predicted results, the modeling of time and space becomes independent at the current step, i.e., $p^*(t_{\text{current}}|t_{\text{last}},s_{\text{last}})$ and $p^*(s_{\text{current}}|t_{\text{last}},s_{\text{last}})$, which successfully solves the intractable problem of the conditional density $p^*(s|t)$.
This novel learning paradigm completely removes the constraints of model structure parameterization in  existing solutions, allowing accurate and flexible modeling of STPPs.

The multi-step learning process simulates the generation of the spatio-temporal joint distribution; however, the underlying mechanism of each step is still unclear.  
To further facilitate the learning at each step, we design a spatio-temporal co-attention module to characterize spatio-temporal interactions that contribute to the target joint distribution. 
Specifically, we simultaneously  learn spatial attention and temporal attention to capture their fine-grained interactions adaptively, which characterizes underlying mechanisms of the joint distribution. 
Table~\ref{tbl:compare} compares the advantages of our framework with existing solutions. 
DSTPP can learn spatio-temporal joint distributions without any dependence restrictions. 
As no integrals or Monte Carlo approximations are required, it is flexible and can perform sampling in a closed form.  It can also be utilized to model a variety of STPPs, where events are accompanied with either a vector of real-valued spatial location or a discrete value, e.g., a class label of the location; 
thus  it is broadly applicable in real-world scenarios.
We summarize our contributions as follows:

\begin{itemize}[leftmargin=*]
    \item To the best of our knowledge, we are the first to model STPPs within the diffusion model paradigm. 
    By removing integrals and overcoming structural design limitations in existing solutions, it achieves flexible and accurate modeling of STPPs.
    
    \item We propose a novel spatio-temporal point process model, DSTPP. On the one hand, the diffusion-based approach decomposes the complex spatio-temporal joint distribution into tractable distributions. On the other hand, the elaborated co-attention module captures the spatio-temporal interdependence adaptively. 
    
    \item Extensive experiments demonstrate the superior performance of our approach for modeling STPPs using both synthetic and real-world datasets. 
    Further in-depth analyses validate that our model successfully captures spatio-temporal interactions for different scenarios in an adaptive manner. 
\end{itemize}

\section{Preliminaries}

\begin{table}[t!]
\caption{Comparison of the proposed model with other point process approaches regarding important properties.}\label{tbl:compare}
\begin{threeparttable}
\resizebox{\columnwidth}{!}{
\begin{tabular}{ccccc}
\hline
Model & No Asmp.\tnote{(1)} & No Restr.\tnote{(2)} & Flexible\tnote{(3)} & \begin{tabular}[c]{@{}c@{}}Closed-form\\ sampling\tnote{(4)}\end{tabular} \\ \hline
Hawkes~\cite{hawkes1971point}    &    \textcolor{red}{\ding{55}}   &  \textcolor{red}{\ding{55}}     &  \textcolor{red}{\ding{55}}   &    \textcolor{red}{\ding{55}}    \\
 Self-correcting~\cite{isham1979self}  &  \textcolor{red}{\ding{55}} &  \textcolor{red}{\ding{55}}      &    \textcolor{red}{\ding{55}}       &   \textcolor{red}{\ding{55}}     \\
KDE~\cite{baddeley2007spatial}  &  -  &   -   &   \textcolor{red}{\ding{55}}  &  \textcolor{green}{\ding{51} }     \\
 CNF~\cite{chen2020neural}  &  -  &  -   &  \textcolor{red}{\ding{55}}  &  \textcolor{green}{\ding{51} }   \\
 ST Hawkes~\cite{reinhart2018review}  &   \textcolor{red}{\ding{55}}  &  \textcolor{red}{\ding{55}}   &  \textcolor{red}{\ding{55}}    &   \textcolor{red}{\ding{55}}  \\
 RMTPP~\cite{du2016recurrent} &  \textcolor{red}{\ding{55}}  &  \textcolor{red}{\ding{55}} &   \textcolor{green}{\ding{51} }    &   \textcolor{red}{\ding{55}} \\ 
 NHP~\cite{mei2017neural} &   \textcolor{red}{\ding{55}} &   \textcolor{red}{\ding{55}} &   \textcolor{green}{\ding{51} }    &   \textcolor{red}{\ding{55}} \\ 
 THP~\cite{zuo2020transformer}  &  \textcolor{red}{\ding{55}} &  \textcolor{red}{\ding{55}} &   \textcolor{green}{\ding{51} }    &   \textcolor{red}{\ding{55}} \\ 
SNAP~\cite{zhang2020self}  &  \textcolor{red}{\ding{55}} &   \textcolor{red}{\ding{55}} &   \textcolor{green}{\ding{51} }    &   \textcolor{red}{\ding{55}} \\ 
LogNormMix~\cite{shchur2019intensity}  &   \textcolor{red}{\ding{55}} &  \textcolor{red}{\ding{55} }  & \textcolor{red}{\ding{55}}   &    \textcolor{green}{\ding{51} }    \\ 
NJSDE~\cite{jia2019neural} &    \textcolor{red}{\ding{55}} & \textcolor{red}{\ding{55}} &   \textcolor{green}{\ding{51} }    &   \textcolor{red}{\ding{55}} \\
 Neural STPP~\cite{chen2020neural} &    \textcolor{green}{\ding{51}} & \textcolor{red}{\ding{55}} &   \textcolor{green}{\ding{51} }    &   \textcolor{red}{\ding{55}} \\ 
DeepSTPP~\cite{zhou2022neural} &    \textcolor{red}{\ding{55}} & \textcolor{red}{\ding{55}} &   \textcolor{green}{\ding{51}}    &   \textcolor{red}{\ding{55}} \\ \hline
DSTPP (ours)    &   \textcolor{green}{\ding{51}}   &   \textcolor{green}{\ding{51} }    &  \textcolor{green}{\ding{51} }        &    \textcolor{green}{\ding{51} }   \\   \hline  

\end{tabular}}
\begin{tablenotes}
        \footnotesize
        \item[(1)] Without assumptions of conditional spatio-temporal independence. 
        \item[(2)] Without dependence restrictions between time and space.
        \item[(3)] Any powerful network architecture can be employed during the calculation.
        \item[(4)] Sampling without any approximation.
      \end{tablenotes}
    \end{threeparttable}
\end{table}

\subsection{Spatio-temporal Point Process}

A spatio-temporal point process is a stochastic process composed of  events with time and space that occur over a domain~\cite{moller2003statistical}.  These spatio-temporal events are described in continuous time with spatial information. The spatial domain of the event can be recorded in different ways. For example, in earthquakes, it is usually recorded  as longitude-latitude coordinates in continuous space. It can also be associated with discrete labels, such as the neighborhoods of crime events. 
Let $x_i=(t_i,s_i)$ denotes the $i_{th}$ spatio-temporal event written as the pair of occurrence time $t\in\mathbb{T}$ and location $s\in\mathbb{S}$, where $\mathbb{T}\times\mathbb{S}\in\mathbb{R}\times\mathbb{R}^d$. Then a spatio-temporal point process can be defined as a sequence $S=\{x_1,x_2,...,x_L\}$, and the number of events $L$ is also stochastic. 
Let  $H_t = \{x_i | t_i < t, x_i \in S\}$ denote the event history before time $t$, 
modeling STPPs is concerned with parameterizing  the conditional probability density function $p(t,s|H_t)$, which denotes the conditional probability density of the next event happening at time $t$ and space $s$ given the history $H_t$.

\textbf{Discussion on shortcomings.} In existing methods for STPPs, given the event history, space and time are assumed to be conditionally independent~\cite{reinhart2018review,zhou2022neural,du2016recurrent,mei2017neural,zuo2020transformer,lin2022exploring} or unilaterally dependent~\cite{daley2003introduction,chen2020neural} i.e., the space is dependent on the time by $p(x|t)$. 
These dependence restrictions destroy the model's predictive performance on entangled space and time interactions conditioned on  history. Besides, most approaches require integration operations when calculating the likelihood, or limit intensity functions to integrable forms, leading to a trade-off between accuracy and efficiency. 
We compare the shortcomings of existing approaches in Table~\ref{tbl:compare}\footnote{TPP models can be used for STPPs where the space acts as the marker.}, which motivate us to design a more flexible and effective model.

\subsection{Denoising Diffusion Probabilistic Models}
Diffusion models~\cite{ho2020denoising} generate samples by learning a distribution that approximates a data distribution. The distribution is learned by a gradual reverse process of adding noise, which recovers the actual value starting from Gaussian noise. At each step of the denoising process, the model learns to predict a slightly less noisy value. 

Let $x^0\thicksim q(x^0)$ denote  a multivariate variable from specific input space $X\in\mathbb{R}^D$, and we consider a probability density function $p_\theta(x^0)$, which aims to approximate $q(x^0)$. Diffusion models are latent variable models, which are defined by two processes: the forward diffusion process and the reverse denoising process. Let $X^k$ for $t=1,2,..., K$ denote a sequence of latent variables of dimension $\in\mathbb{R}^D$, the forward diffusion process is defined by a Markov chain:

\begin{equation}
    q(x^{1:K}|x^0) = \prod_{k=1}^Kq(x^k|x^{k-1})\ \ ,
\end{equation}

\noindent where  $q(x^k|x^{k-1}) \coloneqq \mathcal{N}(x^k;\sqrt{1-\beta_k}x^k$ and $\beta_k\bm{I}), \beta_1,...,\beta_K  \in (0,1)$ is a given increasing variance schedule, representing a particular noise level. 
$x^k$ can be sampled in a closed form as $q(x^k|x^0)=(x^k;\sqrt{\overline{\alpha}_k}x^0, (1-\overline{\alpha}_k)\bm{I})$, 
where 
$\alpha_k \coloneqq 1-\beta_k$ and $\overline{\alpha}_k = \prod_{k=1}^K\alpha_k$. 
Then a noisy observation at the $k_{th}$ step can be expressed as $x^k=\sqrt{\overline{\alpha}_k}x^0+(1-\overline{\alpha}_k)\epsilon$, where $\epsilon \thicksim \mathcal{N}(0,\bm{I})$ and  $x^0$ is the clean observation.  

On the contrary, the reverse denoising process recovers $x^0$ starting from $x^K$, where $x^K\thicksim \mathcal{N}(x^K;0,\bm{I})$. It is defined by the following Markov chain with learned Gaussian transitions:

\begin{equation}
    \begin{aligned}
    &p_\theta(x^{0:K}) \coloneqq p(x^K)\prod_{k=1}^Kp_\theta(x^{k-1}|x^k)\label{eq:sample_o}\ \ , \\
    &p_\theta(x^{k-1}|x^k) \coloneqq \mathcal{N}(x^{k-1};\mu_\theta(x^k,k),\sigma_\theta(x^k,k)\bm{I})\ \ , 
    \end{aligned}
\end{equation}

\noindent $p_\theta(x^{k-1}|x^k)$ aims to remove the Gaussian noise added in the forward diffusion process. The parameter $\theta$ can be optimized by minimizing the negative log-likelihood via a variational bound:

\begin{equation}\label{eq:vbl}
   \min_{\theta}\mathbb{E}_{q(x^0)} \leq  \min_{\theta}\mathbb{E}_{q(x^{0:K})}[-\text{log}p(x^K)-\sum_{k=1}^K\text{log}\frac{p_\theta(x^{k-1}|x^k)}{q(x^k|x^{k-1})}]\ \ . 
\end{equation}
\noindent Ho et al.~\cite{ho2020denoising} show that the denoising parameterization can be trained by the simplified objective: 

\begin{equation}
    \mathcal{E}_{x^0\thicksim q(x^0),\epsilon\thicksim\mathcal{N}(0,\bm{I})}[\|\epsilon-\epsilon_\theta(x_k,k)\|^2]\ \ , 
\end{equation}

\noindent where $x^k = \sqrt{\overline{\alpha}_k}x^0+(1-\overline{\alpha}_k)\epsilon$. 
$\epsilon_\theta$ 
needs to estimate Gaussian noise added to the input $x^k$, which is trained by MSE loss between the real noise and predicted noise. 
Therefore, $\epsilon_\theta$ acts as the denoising network to transform $x^k$ to $x^{k-1}$. 
Once trained, we can sample $x^{k-1}$ from $p_\theta(x^{k-1}|x^k)$  and progressively obtain $x^0$ according to Equation~\eqref{eq:sample_o}.

\section{Spatio-temporal Diffusion Point Processes}

\begin{figure}[t]
    \centering
    \includegraphics[width=0.8\linewidth]{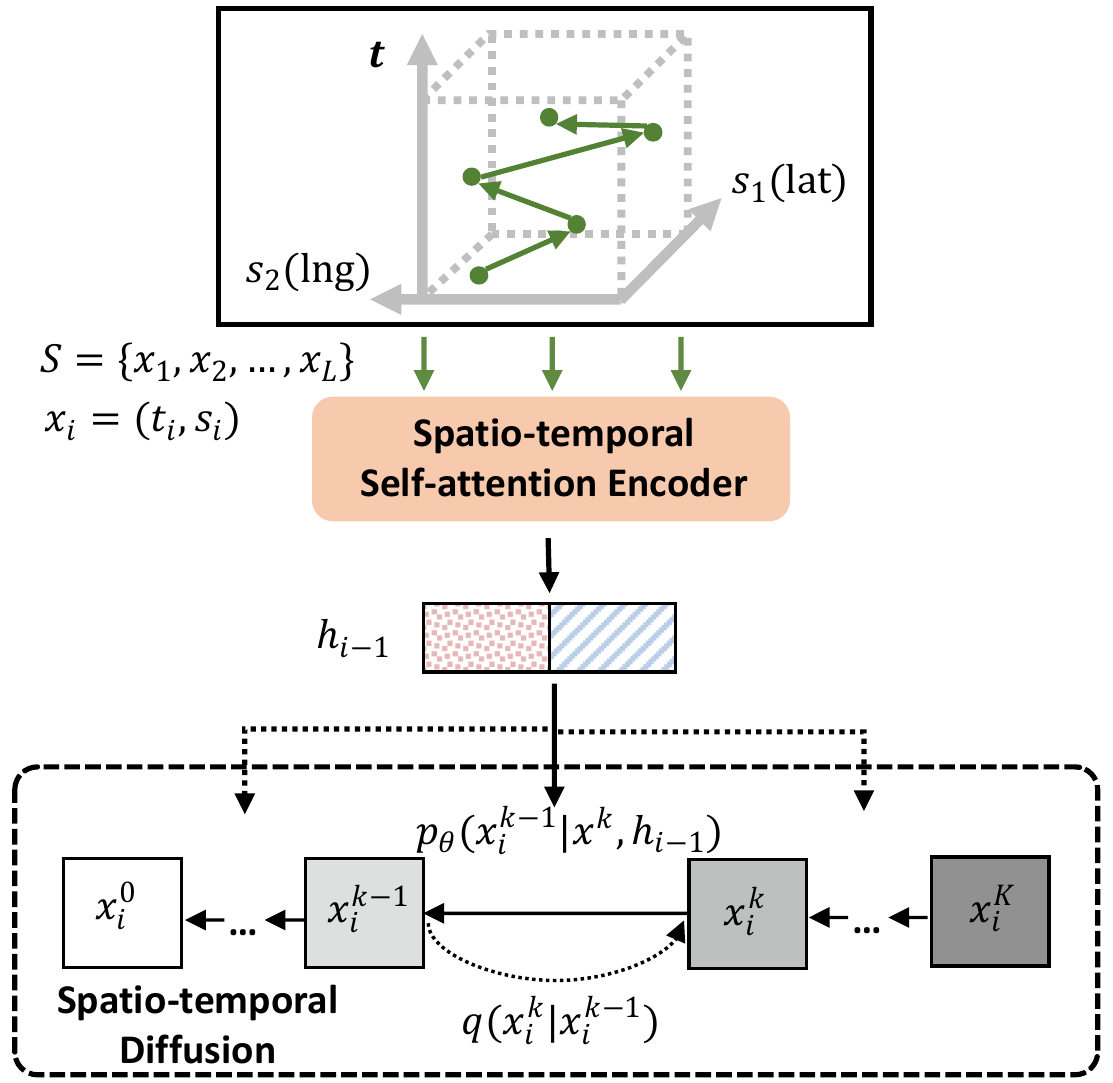}
    \caption{The overview of the proposed DSTPP framework. }
    \label{fig:model}
\end{figure}

Figure~\ref{fig:model} illustrates the overall framework of DSTPP, which consists of two key modules, the spatio-temporal self-attention encoder, and the spatio-temporal diffusion model. 
The spatio-temporal encoder learns an effective representation of the event history, then it acts as the condition to support the spatio-temporal denoising diffusion process. 
We first present the spatio-temporal encoder in Section~\ref{sec:method_0}. Then we formulate the learning of the spatio-temporal joint distribution as a denoising diffusion process, and introduce the diffusion process and inverse denoising process in Section~\ref{sec:method_1}. We describe how to train this model and perform sampling in Section~\ref{sec:method_2}. Finally, We demonstrate the detailed architecture of the denoising network parametrization in Section~\ref{sec:method_3}.

\renewcommand{\algorithmicrequire}{\textbf{Input:}}

\begin{algorithm}[t]
\caption{Training for each spatio-temporal event $x_i=(\tau_i,s_i)$}
\begin{algorithmic}[0]
\REQUIRE
$h_{i-1}$
\renewcommand{\algorithmicrequire}{\textbf{Repeat:}}
\REQUIRE
    \state $x_i^0 \thicksim q(x_i^0)$, 

    \state $k\thicksim \text{Uniform}(1, 2, ..., K)$
    
    \state $\epsilon\thicksim\mathcal{N}(0,I)$

    \state Take gradient descent step on
    $$\nabla_{\phi, \theta}\|\epsilon-\epsilon_\theta(\sqrt{\overline{\alpha}_k}x_i^0+\sqrt{1-\overline{\alpha}_k}\epsilon,h_{i-1},k) \|^2$$
\renewcommand{\algorithmicrequire}{\textbf{Until:}}
\REQUIRE Converged
\label{alg:train}
\end{algorithmic}
\end{algorithm}

\subsection{Spatio-temporal Encoder}\label{sec:method_0}

To model the spatio-temporal dynamics of events and obtain effective sequence representations, we design a self-attention-based spatio-temporal encoder. The input of the encoder is made up of events $x = (t,s)$. To obtain a unique representation for each event, we use two embedding layers for the time and space separately.  For the space $s\in\mathbb{R}^n$, we utilize a linear embedding layer; for the timestamp, we apply a positional encoding method following~\cite{zuo2020transformer}: 

\begin{gather}
    [e_t]_j = \begin{cases}
    cos(t/10000^{\frac{j-1}{M}})  & \text{if } j \text{ is odd} \\
    sin(t/10000^{\frac{j-1}{M}}) & \text{if } j \text{ is even}\ \ , 
    \end{cases}
\end{gather}

\noindent where $e_t$ denotes the temporal embedding and $M$ is the embedding dimension.
For the spatial domain, we use linear projection to convert continuous or discrete space into embeddings as follows:
\begin{gather}
    e_s = W_es
\end{gather}

\noindent where $W_e$ contains learnable parameters. 

We use $W_e\in\mathcal{R}^{M\times D}$ if the space $s$ is defined in the continuous domain $\mathbb{R}^D, D\in\{1,2,3\}$. 
We use $W_e\in\mathcal{R}^{M\times N}$ if the spatial information is associated with  discrete locations represented by one-hot ID encoding $s\in\mathbb{R}^{N}$, where $N$ is the number of discrete locations. In this way, we obtain real-value vectors $e_s$ for both continuous and discrete spatial domains. 
For each event $x=(t,s)$, we obtain the spatio-temporal embedding $e_{st}$ by adding the positional encoding $e_t$ and spatial embedding $e_s$. The embedding of the $S = \{(t_i, s_i)\}_{i=1}^L$ is then specified by $E_{st} = \{e_{st,1}, e_{st,2}, ..., e_{st,L}\} \in \mathbb{R}^{L \times M}$, where $e_{st,i} = e_{s,i} + e_{t,i}$. In the meantime, we also keep the temporal embedding $E_t=\{e_{t,1},e_{t,2},...,e_{t,L}\}$ and spatial embedding $E_s=\{e_{s,1},e_{s,2},...,e_{s,L}\}$, respectively, with the goal of capturing characteristics of different aspects.
If only spatio-temporal representation is available, the model may fail when dealing with cases where the temporal and spatial domains are not entangled. With learned representations from different aspects, we did not simply sum them together. Instead, we concatenate them and enable the model to leverage representations adaptively.

After the initial spatial embedding and temporal encoding layers, we pass $E_{st}$, $E_s$, and $E_t$ through three self-attention modules. Specifically, the scaled dot-product attention~\cite{vaswani2017attention} is defined as:

\begin{equation}
    \begin{aligned}
    &\text{Attention}(Q,K,V) = \text{Softmax}(\frac{QK^T}{\sqrt{d}})\ \ , \\
    &S = \text{Attention}(Q,K,V) V\ \ ,
    \end{aligned}
\end{equation}

\noindent where $Q, K,$ and $V$ represent queries, keys, and values. In our case, the self-attention operation takes the embedding $E$ as input, and then converts it into three matrices by linear projections:
\begin{gather}
    Q=EW^Q,  K=EW^K,  V=EW^V,
\end{gather}

\noindent where $W^Q, W^K$, and $W^V$  are weights of linear projections. Finally, we use a position-wise feed-forward network to transform the attention output $S$ into the hidden representation $h(t)$. 

For three embeddings $E_s, E_t$ and $E_{st}$ containing information of different aspects, we all
employ the above self-attentive operation to generate hidden spatial representation $h_s(t)$, temporal representation $h_t(t)$, and spatial-temporal representation $h_{st}(t)$. As a result, the hidden representation $h_{i-1}$ in Figure~\ref{fig:model} is a collection of the three representations.

\subsection{Spatio-temporal Diffusion and Denoising Processes}\label{sec:method_1}

Conditioned on the hidden representation $h_{i-1}$ generated by the encoder, we aim to learn a model of the spatio-temporal joint distribution of the future event. 
The learning of such distribution is built on the diffusion model~\cite{ho2020denoising}, and the values of space and time are diffused and denoised at each event. Specifically, for each event $x_i=(\tau_i,s_i)$ in the sequence, where $\tau_i$ denotes the time interval since the last event, we model the diffusion process as a Markov process over the spatial and temporal domains as $(x_i^0,x_i^1,...,x_i^K)$, where $K$ is the number of diffusion steps. From $x_i^0$ to $x_i^K$, we add a little Gaussian noise step by step to the space and time values until they are corrupted into pure Gaussian noise. The process of adding noise is similar to image scenarios, where the noise is applied independently on each pixel~\cite{ho2020denoising}. We diffuse separately on the spatial and temporal domains by the following probabilities:
\begin{equation}
\begin{aligned}
&q_{st}(\bm{x}^k_i|\bm{x}^{k-1}_i) \coloneqq (q(\tau^k_i|\tau_i^{k-1}),q(s_i^k|s_i^{k-1}))\ \ , \\
&q(x^k|x^{k-1}) \coloneqq \mathcal{N}(x^k;\sqrt{1-\beta_k}x^k, \beta_k\bm{I})\ \ , 
\end{aligned}
\end{equation}

\noindent where $\alpha_k=1-\beta_k$ and $\overline{\alpha}_k=\prod_{s=1}^k\alpha_k$.

\renewcommand{\algorithmicrequire}{\textbf{Input:}}
\let\oldReturn\Return

\begin{algorithm}[t]
\caption{Sampling $s_i^0$ and $\tau_i^0$}
\begin{algorithmic}[0]

\REQUIRE
Noise $s_i^K\thicksim \mathcal{N}(0,I)$, $\tau_i^K\thicksim \mathcal{N}(0,I)$ and $h_{i-1}$
\FOR{k = K to 1}
\STATE {$z_s\thicksim\mathcal{N}(0,I), z_t\thicksim\mathcal{N}(0,I)$ if k>1 else $z_s=0, z_t=0$}
\STATE {$s_i^{k-1}=\frac{1}{\sqrt{\alpha_k}}(s_i^k-\frac{\beta_k}{\sqrt{1-\overline{\alpha}_k}}\epsilon_\theta(s_i^k, \tau_i^k, h_{i-1}, k)) + \sqrt{\beta_k}z_s$}
\STATE {$\tau_i^{k-1}=\frac{1}{\sqrt{\alpha_k}}(\tau_i^k-\frac{\beta_k}{\sqrt{1-\overline{\alpha}_k}}\epsilon_\theta(s_i^k, \tau_i^k, h_{i-1}, k)) + \sqrt{\beta_k}z_t$}
\ENDFOR
\renewcommand{\algorithmicrequire}{\textbf{Return:}}
\REQUIRE
$s_i^0, \tau_i^0$
\end{algorithmic}\label{alg:sample}
\end{algorithm}

On the contrary, we formulate the reconstruction of the point $x_i=(\tau_i,s_i)$ as reverse denoising iterations from $x_i^K$ to $x_i^0$ given the event history. In addition to the history representation $h_{i-1}$, the denoising processes of time and space are also dependent on each other obtained in the previous step. The predicted values of the next step are modeled in a conditionally independent manner, which is formulated as follows:

\begin{equation}\label{eq:st_joint}
    p_\theta(x_i^{k-1}|x_i^k, h_{i-1}) = p_\theta(\tau_i^{k-1}|\tau_i^k, s_i^k, h_{i-1}) p_\theta(s_i^{k-1}|\tau_i^k, s_i^k, h_{i-1})\ \ ,
\end{equation}

\noindent In this way, we manage to disentangle the modeling of spatio-temporal joint distribution into conditionally independent modeling, which enables effective and efficient modeling of the observed spatio-temporal distribution. 
The overall reverse denoising process is formulated as follows:
\begin{gather}\label{eq:denoise}
    p_\theta(x_i^{0:K}|h_{i-1}) \coloneqq p(x_i^K) \prod_{k=1}^Kp_\theta(x_i^{k-1}|x_i^k, h_{i-1})\ \ .
\end{gather}

\noindent For the continuous-space domain, the spatio-temporal distribution can be predicted by Equation~\ref{eq:denoise}. For the discrete-space domain, we add a rounding step at the end of the reverse process, $p_\theta(s_i|s_i^0)$, to convert the real-valued embedding $s_i^0$ to discrete location ID $s_i$.

\subsection{Training and Inference}\label{sec:method_2}

\subsubsection*{\textbf{Training}}

For a spatio-temporal point process, the training should optimize the parameters $\theta$ that maximize the log-likelihood:

\begin{gather}
    \sum_{i=1}^L\text{log}p_\theta(x_i^0|h_{i-1})\ \ ,
\end{gather}

\noindent where $L$ is the number of events in the sequence. Based on a similar derivation in the preliminary section, we train the model by a simplified loss function for the $i_{th}$ event and diffusion step $k$  as follows~\cite{ho2020denoising}:

\begin{gather}
    \mathcal{L} = \mathbb{E}_{x_i^0,\epsilon,k}[\|\epsilon-\epsilon_\theta(\sqrt{\overline{\alpha}_k}x_i^0+\sqrt{1-\overline{\alpha}_k}\epsilon,h_{i-1},k) \|^2]\ \ ,
\end{gather}

\noindent where $\epsilon\thicksim\mathcal{N}(0,I)$. Samples at each diffusion step k for each event are included in the training set. We train the overall framework consisting of ST encoder and ST diffusion in an end-to-end manner.   The pseudocode of the training procedure is shown in Algorithm~\ref{alg:train}.

\subsubsection*{\textbf{Inference}}
To predict future spatio-temporal events with trained DSTPP. We first obtain the hidden representation $h_i$ by employing the spatio-temporal self-attention encoder given past $i-1$ events.  Then, we can predict the next event starting from Gaussian noise $s_i^K, \tau_i^K\thicksim \mathcal{N}(0,I)$ conditioned on $h_i$.
Specifically, the reconstruction of $x_i^0$ from $x_i^K=(s_i^K, \tau_i^K)$ is formulated as follows:

\begin{equation}\label{eq:Inference}
    \begin{aligned}
        s_i^{k-1}=\frac{1}{\sqrt{\alpha_k}}(s_i^k-\frac{\beta_k}{\sqrt{1-\overline{\alpha}_k}}\epsilon_\theta(x_i^k, h_{i-1}, k)) + \sqrt{\beta_k}z_s \ \ , \\
        \tau_i^{k-1}=\frac{1}{\sqrt{\alpha_k}}(\tau_i^k-\frac{\beta_k}{\sqrt{1-\overline{\alpha}_k}}\epsilon_\theta(x_i^k, h_{i-1}, k)) + \sqrt{\beta_k}z_t\ \ ,
    \end{aligned}
\end{equation}

\noindent where $z_s$ and $z_t$ are both stochastic variables sampled from a standard Gaussian distribution. $\epsilon_\theta$ is the trained reverse denoising network, which takes in the previous denoising result $x_i^k$, the hidden representation of the sequence history $h_{i-1}$ and the diffusion step $k$. Algorithm~\ref{alg:sample} presents the pseudocode of the sampling procedure.

\begin{figure}[t]
    \centering
    \includegraphics[width=0.99\linewidth]{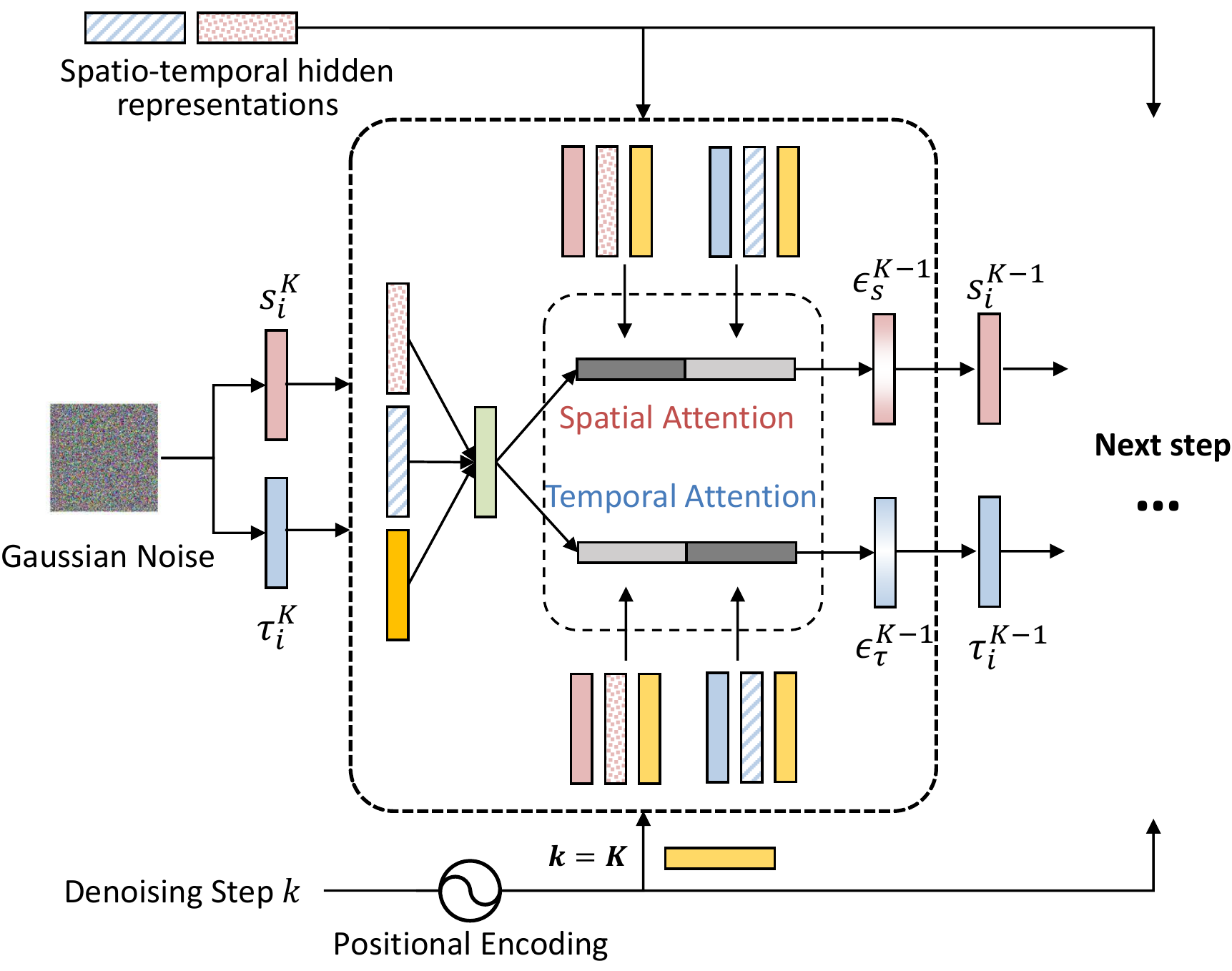}
    \caption{Network architecture of the spatio-temporal co-attention mechanism. Each step in the denoising process shares the same network structure, with spatio-temporal hidden representations as conditions. }
    \label{fig:CA}
\end{figure}

\subsection{Co-attention Denoising Network}\label{sec:method_3}
We design a co-attention denoising network to capture the interdependence between spatial and temporal domains, which facilitates the learning of spatio-temporal joint distributions.  Specifically,  it performs spatial and temporal attention simultaneously at each denoising step to capture fine-grained interactions.  Figure~\ref{fig:CA} illustrates the detailed network architecture. Each step of the denoising process shares the same structure, which takes in the previously predicted values $s_i^{k+1}$ and $\tau_i^{k+1}$, and the denoising step $k$ with positional encoding. Meanwhile,  the network also integrates
the hidden representation $h_{i-1}$ to achieve conditional denoising. 

Temporal attention aims to generate a  context vector by attending to certain parts of the temporal input  and certain parts of the spatial input, and so does spatial attention. We calculate the mutual attention weights, i.e., $\alpha_s$  and $\alpha_t$, for space and time  based on the condition $h_{i-1}$ and current denoising step $k$ as follows:

\begin{equation}
    \begin{aligned}
    e_k &=  \text{SinusoidalPosEmb}(k)\ \ , \\
    \alpha_s &= \text{Softmax}(W_{sa}\text{Concat}(h_{i-1},e_{k})+b_{sa})\ \ ,  \\
    \alpha_t &= \text{Softmax}(W_{ta}\text{Concat}(h_{i-1},e_{k})+b_{ta})\ \ ,
    \end{aligned}
\end{equation}

\noindent  where $W_{sa}, W_{ta}, b_{sa}, b_{ta}$ are learnable parameters. $\alpha_s$ and $\alpha_t$ measure the mutual dependence between time and space, which are influenced by the event history and current denoising step.

Then we integrate the spatio-temporal condition $h_{i-1}=\{h_{s,i-1}$, $h_{t,i-1}\}$ into previously predicted values $s_i^{k+1}$ and $\tau_i^{k+1}$ by feed-forward neural networks, and each layer is formulated as follows:

\begin{equation}
    \begin{aligned}
    x_{s,i} = \sigma (W_ss_i^{k+1}+b_s + W_{sh}h_{s,i-1}+b_{sh} + e_k) \ \ , \\
    x_{t, i} = \sigma (W_t\tau_i^{k+1}+b_t + W_{th}h_{t,i-1}+b_{th} + e_k)\ \ ,
    \end{aligned}
\end{equation}

\noindent where $W_s\in \mathbb{R}^{M\times D}, W_t\in \mathbb{R}^{M\times 1}, W_{sh}, W_{th} \in \mathbb{R}^{M\times M}$, and $b_s, b_t,  b_{sh}$,  $b_{th} \in \mathbb{R}^{M\times 1}$ are learnable parameters of the linear projection, and $\sigma$ denotes the ReLU activation function. Finally, the outputs of spatial attention and temporal attention are calculated as follows:

\begin{equation}
    \begin{aligned}
    &x_i = [x_{s,i}, x_{t,i}] \ \ ,\\
    &\epsilon_{s,i}^k = \sum \alpha_sx_i\ \ ,  \epsilon_{t,i}^k = \sum \alpha_tx_i, 
    \end{aligned}
\end{equation}

\noindent where $\epsilon_{s,i}^k$ and $\epsilon_{t,i}^k$ are the predicted noise at step $k$ for the $i_{th}$ event.  We can obtain the predicted values  $s_i^{k}$ and $\tau_i^{k}$ at step $k$ according to Equation \eqref{eq:Inference}.  Then the predicted values  $s_i^{k}$ and $\tau_i^{k}$ are fed into the denoising network again to iteratively predict the results towards the clean values of space and time. 
In this way, the interdependence between time and space is captured adaptively and dynamically,  facilitating the learning of the spatio-temporal joint distribution.

\section{Experiments}

In this section, we perform experiments to answer the following research questions:

\begin{itemize}[leftmargin=*]
    \item  \textbf{RQ1:} How does the proposed model perform compared with existing baseline approaches?
    \item \textbf{RQ2:} Is the joint modeling of spatial and temporal dimensions effective for STPPs, and what's the spatio-temporal interdependence like during the denoising process?
    \item \textbf{RQ3:} How does the total number of diffusion steps affect the performance?
    \item \textbf{RQ4:} How to gain a deeper understanding of the reverse denoising diffusion process? 
\end{itemize}

\subsection{Experimental Setup}
\subsubsection{\textbf{Datasets}}
We perform extensive experiments on synthetic datasets and real-world datasets in the STPP literature. All datasets are obtained from open sources, which contain up to thousands of spatio-temporal events.
Varying across a wide range of fields, we use one synthetic dataset and three real-world datasets, including earthquakes in Japan, COVID-19 spread, bike sharing in New York City, and simulated Hawkes Gaussian Mixture Model process~\cite{chen2020neural}.  
Besides, we use a real-world dataset, Atlanta Crime Data, the spatial locations of which are discrete neighborhoods.  
We briefly introduce them here, and further details can be found in Appendix~\ref{sup:data}.

\textbf{(1) Earthquakes.} Earthquakes in Japan with a magnitude of at least 2.5 from 1990 to 2020 recorded by the U.S. Geological Survey\footnote{https://earthquake.usgs.gov/earthquakes/search/}. 
\textbf{(2) COVID-19.}  Publicly released by The New York Times (2020), which records daily infected cases of COVID-19 in New Jersey state\footnote{https://github.com/nytimes/covid-19-data}. We aggregate the data at the county level.  
\textbf{(3) Citibike.} Bike sharing in New York City collected by a bike sharing service. The start of each trip is considered as an event. 
\textbf{(4)HawkesGMM\footnote{https://github.com/facebookresearch/neural\_stpp/blob/main/toy\_datasets.py}.} This synthetic data uses Gaussian Mixture Model to generate spatial locations. Events are sampled from a multivariate Hawkes process. 
\textbf{(6) Crime~\footnote{http://www.atlantapd.org/i-want-to/crime-data-downloads}.} It is provided by the Atlanta Police Department, recording robbery crime events. Each event is associated with the time and the neighborhood. 

\begin{table*}[th]
\caption{Performance evaluation for negative log-likelihood per event on test data.  $\downarrow$ means lower is better. Bold denotes the best results and \underline{underline} denotes the second-best results.}\label{tbl:nll}
\begin{threeparttable}
\begin{tabular}{c|cc|cc|cc|cc}
\hline
& \multicolumn{2}{c|}{Earthquake} & \multicolumn{2}{c|}{COVID-19} & \multicolumn{2}{c|}{Citibike} & \multicolumn{2}{c}{HawkesGMM} \\ \cline{2-9} 
Model            & Spatial  $\downarrow$  & Temporal $\downarrow$  & Spatial  $\downarrow$ & Temporal $\downarrow$  & Spatial  $\downarrow$  & Temporal $\downarrow$  & Spatial   $\downarrow$ & Temporal  $\downarrow$  \\ \hline
Conditional KDE  &    2.21{\scriptsize $\pm$0.105}     &   -\tnote{(1)}  &   2.31{\scriptsize $\pm$0.084}       &   -  &   2.74{\scriptsize $\pm$0.001}     &   -    &  \underline{0.236{\scriptsize $\pm$0.001}}       &      -       \\
CNF              &     1.35{\scriptsize $\pm$0.000}     & -   &   2.05{\scriptsize $\pm$0.014}       &   - &   2.15{\scriptsize $\pm$0.000}     &  -    &  0.427{\scriptsize $\pm$0.002}       &  -      \\
TVCNF &        1.34{\scriptsize $\pm$0.008}     &  -   &   2.04{\scriptsize $\pm$0.004}       &   -   &   2.19{\scriptsize $\pm$0.025}     &  -  &  0.431{\scriptsize $\pm$0.008}       &  -       \\
Possion          &    -       &    -0.146{\scriptsize $\pm$0.000}        &     -      &     -0.876{\scriptsize $\pm$0.021}       &    -       &   -0.626{\scriptsize $\pm$0.000}         &     -      &      1.34{\scriptsize $\pm$0.000}             \\
Hawkes           &      -     &      -0.514{\scriptsize $\pm$0.000}        &    -       &     -2.06{\scriptsize $\pm$0.000}       &    -       &   -1.06{\scriptsize $\pm$0.001}         &     -      &      0.880{\scriptsize $\pm$0.000}             \\
Self-correcting  &     -      &      13.8{\scriptsize $\pm$0.533}        &    -       &     7.13{\scriptsize $\pm$0.062}       &     -      &   7.11{\scriptsize $\pm$0.010}         &   -        &      4.59{\scriptsize $\pm$0.135}             \\
RMTPP              &  -   &    0.0930{\scriptsize $\pm$0.051}     &    -      &     -1.30{\scriptsize $\pm$0.022}       &     -      &      1.24{\scriptsize $\pm$0.001}    &    -      &       1.52{\scriptsize $\pm$0.002}      \\
NHP             &  -   &   -0.676{\scriptsize $\pm$0.001}   &   -  &  -2.30{\scriptsize $\pm$0.001}    &    -  &  -1.14{\scriptsize $\pm$0.001}    &    -   &    0.580{\scriptsize $\pm$0.000}    \\   
THP              &  -   &   \underline{-0.976{\scriptsize $\pm$0.011}}   &   -  &  -2.12{\scriptsize $\pm$0.002}    &    -  &  \underline{-1.49{\scriptsize $\pm$0.003}}    &    -   &    -0.402{\scriptsize $\pm$0.001}    \\   
SAHP            &  -   &   -0.229{\scriptsize $\pm$0.007}   &   -  &  -1.37{\scriptsize $\pm$0.118}    &    -  &  -1.02{\scriptsize $\pm$0.067}    &    -   &    \underline{-1.25{\scriptsize $\pm$0.136}}    \\   
LogNormMix   &    -       &     -0.341{\scriptsize $\pm$0.071}      &     -     &   -2.01{\scriptsize $\pm$0.025}        &     -      &  -1.06{\scriptsize $\pm$0.005}         &    -     &     0.630{\scriptsize $\pm$0.004}         \\ \hline
NJSDE      &       1.65{\scriptsize $\pm$0.012}    &  0.0950{\scriptsize $\pm$0.203}    &  2.21{\scriptsize $\pm$0.005}     &   -1.82{\scriptsize $\pm$0.002}      &   2.63{\scriptsize $\pm$0.001}   &    -0.804{\scriptsize $\pm$0.059}  &   0.395{\scriptsize $\pm$0.001}   &   1.77{\scriptsize $\pm$0.030}         \\ 
NSTPP      &   \underline{0.885{\scriptsize $\pm$0.037}}      &   -0.623{\scriptsize $\pm$0.004}          &      1.90{\scriptsize $\pm$0.017}         &  \underline{-2.25{\scriptsize $\pm$0.002}} &     2.38{\scriptsize $\pm$0.053}       &   -1.09{\scriptsize $\pm$0.004}         &   0.285{\scriptsize $\pm$0.011}        &       0.824{\scriptsize $\pm$0.005}        \\ 
DeepSTPP        &   4.92{\scriptsize $\pm$0.007}  &   -0.174{\scriptsize $\pm$0.001}      &   \underline{0.361{\scriptsize $\pm$0.01}} &   -1.09{\scriptsize $\pm$0.01}    &   \textbf{-4.94{\scriptsize $\pm$0.016}}  &   -1.13{\scriptsize $\pm$0.002}    &    0.519{\scriptsize $\pm$0.001}   &     0.322{\scriptsize $\pm$0.002}   \\  \hline
DSTPP (ours)    &  \textbf{0.413{\scriptsize $\pm$0.006}}  &   \textbf{-1.10{\scriptsize $\pm$0.020}}      &   \textbf{0.350{\scriptsize $\pm$0.029}} &   \textbf{-2.66{\scriptsize $\pm$0.003}}    &   \underline{0.529{\scriptsize $\pm$0.011}}  &   \textbf{-2.43{\scriptsize $\pm$0.010}}    &    \textbf{0.200{\scriptsize $\pm$0.047}}   &     \textbf{-1.63{\scriptsize $\pm$0.002}}   \\  \hline
\end{tabular}
\begin{tablenotes}
        \footnotesize
        \item[(1)]  \textbf{Spatial baselines and temporal baselines can be combined freely for modeling spatio-temporal domains.}
      \end{tablenotes}
    \end{threeparttable}
\end{table*}
\begin{table*}[th]
\caption{Performance evaluation for predicting both time and space of the next event.
 We use Euclidean distance to measure the prediction error of the spatial domain and use RMSE between real intervals and predicted intervals for time prediction. }\label{tbl:pred}
 \vspace{-3mm}
\begin{tabular}{c|cc|cc|cc|cc}
\hline
& \multicolumn{2}{c|}{Earthquake} & \multicolumn{2}{c|}{COVID-19} & \multicolumn{2}{c|}{Citibike} & \multicolumn{2}{c}{HawkesGMM} \\ \cline{2-9} 
Model            & Spatial $\downarrow$   & Temporal $\downarrow$  & Spatial $\downarrow$   & Temporal $\downarrow$  & Spatial $\downarrow$   & Temporal  $\downarrow$ & Spatial  $\downarrow$  & Temporal $\downarrow$   \\ \hline
Conditional KDE  &  11.3{\scriptsize $\pm$0.658}    &  - &    0.688{\scriptsize $\pm$0.047}   & -     &    0.718{\scriptsize $\pm$0.001}   &    -  &  1.54{\scriptsize $\pm$0.006}    &    -    \\
CNF              &   8.48{\scriptsize $\pm$0.054}    &  -   &    \underline{0.559{\scriptsize $\pm$0.000}}   &      -  &    0.722{\scriptsize $\pm$0.000}   &  -   &   71663{\scriptsize $\pm$60516}    &    -    \\
TVCNF &     8.11{\scriptsize $\pm$0.001}    &  -  &    0.560{\scriptsize $\pm$0.000}   &     -  &    0.705{\scriptsize $\pm$0.000}   &  -  &   2.03{\scriptsize $\pm$0.000}    &    -   \\
Possion          &      -     &   0.631{\scriptsize $\pm$0.017}   &    -        &  0.463{\scriptsize $\pm$0.021}    &      -      &  0.438{\scriptsize $\pm$0.001}    &    -   &    2.81{\scriptsize $\pm$0.070}    \\
Hawkes           &     -     &   0.544{\scriptsize $\pm$0.010}   &    -    &  0.672{\scriptsize $\pm$0.088}    &   -    &  0.534{\scriptsize $\pm$0.011}    &      -    &    2.63{\scriptsize $\pm$0.002}    \\
Self-correcting  &   -     &   11.2{\scriptsize $\pm$0.486}   &     -     &  2.83{\scriptsize $\pm$0.141}    &     -      &  10.7{\scriptsize $\pm$0.169}    &   -    &    9.72{\scriptsize $\pm$0.159}    \\
RMTPP              &   -   &   0.424{\scriptsize $\pm$0.009}     &  -      &     1.32{\scriptsize $\pm$0.024}       &  -       &      2.07{\scriptsize $\pm$0.015}    &   -      &       3.38{\scriptsize $\pm$0.012}      \\
NHP               &  -   &   1.86{\scriptsize $\pm$0.023}   &   -  &  2.13{\scriptsize $\pm$0.100}    &    -  &  2.36{\scriptsize $\pm$0.056}    &    -   &    2.82{\scriptsize $\pm$0.028}    \\   
THP              &  -   &   2.44{\scriptsize $\pm$0.021}   &   -  &  0.611{\scriptsize $\pm$0.008}    &    -  &  1.46{\scriptsize $\pm$0.009}    &    -   &    5.35{\scriptsize $\pm$0.002}    \\   
SAHP            &  -   &   0.409{\scriptsize $\pm$0.002}   &   -  &  0.184{\scriptsize $\pm$0.024}    &    -  &  \underline{0.203{\scriptsize $\pm$0.010}}    &    -   &    2.75{\scriptsize $\pm$0.049}    \\   
LogNormMix   &    -       &      0.593{\scriptsize $\pm$0.005}        &      -     &     0.168{\scriptsize $\pm$0.011}         &      -     &         0.350{\scriptsize $\pm$0.013}     &     -      &       2.79{\scriptsize $\pm$0.021}          \\ 
WGAN   &     -      &     0.481{\scriptsize $\pm$0.007}      &      -     &    \underline{0.124{\scriptsize $\pm$0.002}}        &    -       &     0.238{\scriptsize $\pm$0.003}       &      -     &      2.83{\scriptsize $\pm$0.048}        \\  \hline
NJSDE   &    9.98{\scriptsize $\pm$0.024} &   0.465{\scriptsize $\pm$0.009} &  0.641{\scriptsize $\pm$0.009}  &  0.137{\scriptsize $\pm$0.001} &   0.707{\scriptsize $\pm$0.001}    &  0.264{\scriptsize $\pm$0.005}  &   1.62{\scriptsize $\pm$0.003}   &    2.25{\scriptsize $\pm$0.007} \\ 
NSTPP      &   \underline{8.11{\scriptsize $\pm$0.000}}        &  0.547{\scriptsize $\pm$0.010}  &   0.560{\scriptsize $\pm$0.000}   &   0.145{\scriptsize $\pm$0.002}   &    0.705{\scriptsize $\pm$0.000}&  0.355{\scriptsize $\pm$0.013}  &  2.02{\scriptsize $\pm$0.000} &   3.30{\scriptsize $\pm$0.201}   \\ 
DeepSTPP        & 9.20{\scriptsize $\pm$0.000}   &   \textbf{0.341{\scriptsize $\pm$0.000}}  &  0.687{\scriptsize $\pm$0.000} &  0.197{\scriptsize $\pm$0.000}     &  \underline{0.044{\scriptsize $\pm$0.000}}    &   0.234{\scriptsize $\pm$0.000}     & \underline{1.38{\scriptsize $\pm$0.000}}   &  \underline{1.46{\scriptsize $\pm$0.000}}     \\ \hline
DSTPP (ours)   &   \textbf{6.77{\scriptsize $\pm$0.193}}  &  \underline{0.375{\scriptsize $\pm$0.001}}    & \textbf{0.419{\scriptsize $\pm$0.001}} &        \textbf{0.093{\scriptsize $\pm$0.000}}  & \textbf{0.031{\scriptsize $\pm$0.000}}   &  \textbf{0.200{\scriptsize $\pm$0.002}} &  \textbf{1.28{\scriptsize $\pm$0.013}}   &   \textbf{1.07{\scriptsize $\pm$0.009}}  \\ \hline 
\end{tabular}
\end{table*}

\subsubsection{\textbf{Baselines}}
To evaluate the performance of our proposed model, we compare it with commonly-used methods and state-of-the-art models. The baselines can be divided into three groups: spatial baselines, temporal baselines, and spatio-temporal baselines. It is common for previous methods to model the spatial domain and temporal domain separately, so spatial baselines and temporal baselines can be combined freely for STPPs. We summarize the three groups as follows\footnote{Appendix~\ref{sup:baseline} provides more details of the used baselines.}:
\begin{itemize}[leftmargin=*]
    \item \textbf{Spatial baselines:} We use conditional kernel density estimation (Condition KDE)~\cite{chen2018neural}, Continuous normalizing flow (CNF), and Time-varying CNF~\cite{chen2018neural} (TVCNF)~\cite{chen2018neural}. The three methods all model continuous spatial distributions.
    
    \item \textbf{Temporal baselines:} We include commonly used TPP models. Classical TPP models include the Poisson process~\cite{rasmussen2018lecture}, Hawkes Process~\cite{hawkes2018hawkes}, and Self-correcting process~\cite{isham1979self}.  We also incorporate neural TPP models, including Recurrent Marked Temporal Point Process (RMTPP)~\cite{du2016recurrent}, Neural Hawkes Process (NHP)~\cite{mei2017neural}, Transformer Hawkes Process (THP)~\cite{zuo2020transformer}, Self-attentive Hawkes Process (SAHP) \cite{zhang2020self}. Besides, we also compare with intensity-free approaches: Log Normal Mixture model (LogNormMix)~\cite{shchur2019intensity}, and  Wasserstein GAN (WGAN)~\cite{xiao2017wasserstein}.

    \item \textbf{Spatio-temporal baselines.} We include state-of-the-art spatio-temporal baselines, including Neural Jump Stochastic Differential Equations (NJSDE)~\cite{shchur2019intensity}, Neural Spatio-temporal Point Process (NSTPP)~\cite{chen2020neural}, and DeepSTPP~\cite{zhou2022neural}.
\end{itemize}

\subsubsection{\textbf{Evaluation Metrics}} 
We evaluate the performance of models from two perspectives: likelihood comparison and event prediction comparison.  
We use negative log-loglikelihood (NLL) as metrics, and the time and space are evaluated, respectively.  
Although the exact likelihood cannot be obtained, we can write the variational lower bound (VLB) according to Equation~\eqref{eq:vbl} and utilize it as the NLL metric instead. Thus, the performance on exact likelihood is even better than the reported variational lower bound. 
The models' predictive ability for time and space is also important in practical applications~\cite{okawa2019deep}. Since time intervals are real values, we use a common metric, Root Mean Square Error (RMSE), to evaluate time prediction. The spatial location can be defined in $D$-dimensional space, so we use Euclidean distance to measure the spatial prediction error. We refer the readers to Appendix~\ref{sup:metric} for more details of the used evaluation metrics.


\subsection{Overall performance}

Table~\ref{tbl:nll} and Table~\ref{tbl:pred} show the overall performance of models on NLL and prediction, respectively. Figure~\ref{fig:discrete_main} shows the prediction performance of models in discrete-space scenarios.
From these results, we have the following observations:

\begin{itemize}[leftmargin=*]
    \item \textbf{Unreasonable parametric assumptions for point processes destroy the performance severely.} The worst performance of the self-correcting process indicates the assumption that the occurrence of past events inhibits the occurrence of future events, does not match realities. On the contrary, the Hawkes process, which assumes the occurrence of an event increases the probability of event occurrence in the future, 
    outperforms other classical models (Poisson and Self-correcting), with an obvious reduction of temporal NLL. Nevertheless, the self-exciting assumption can still fail when faced with cases where previous events prevent subsequent events.  Therefore, classical models that require certain assumptions, cannot cover all situations with different dynamics. 

\begin{figure}[t]
    \centering
    \includegraphics[width=0.96\linewidth]{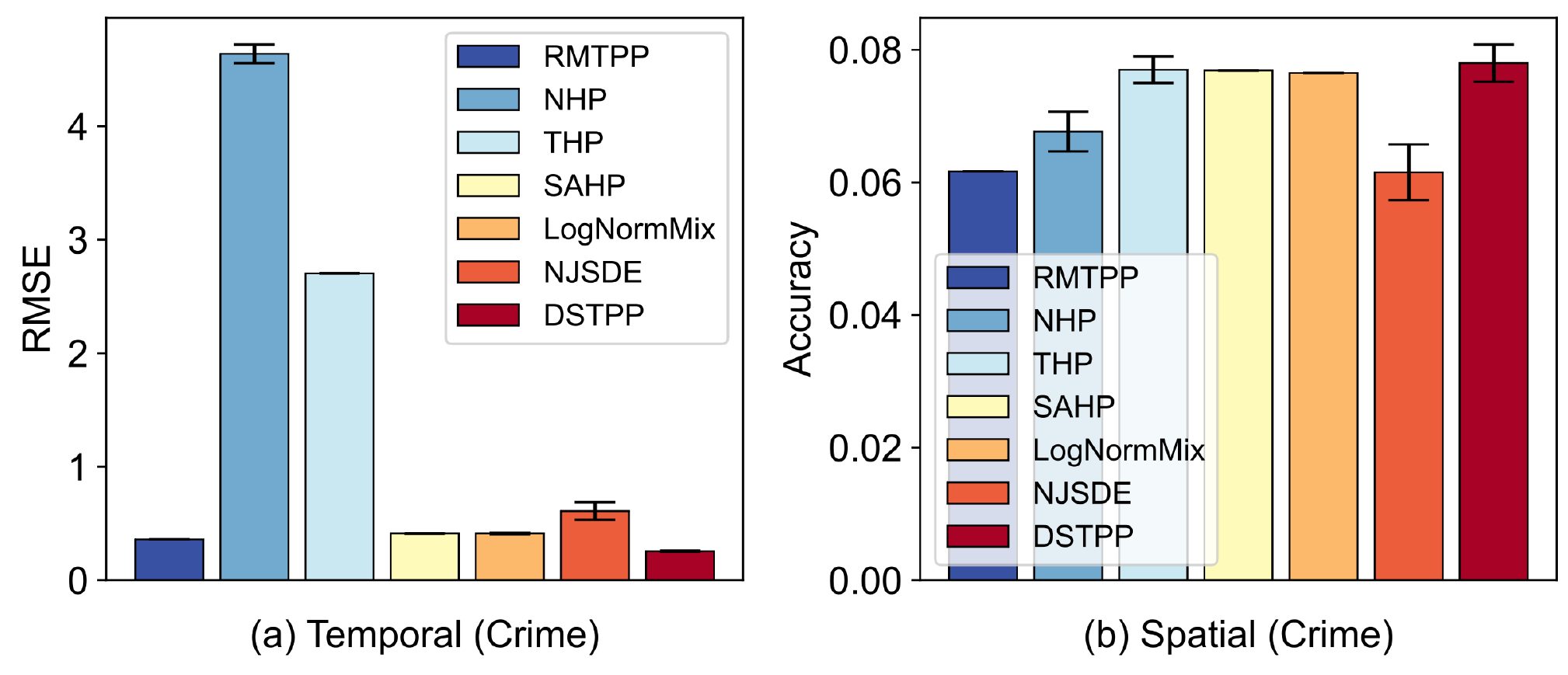}
    \vspace{-5mm}
    \caption{The performance of models on discrete-space datasets for both time and space of the next event.}
    \label{fig:discrete_main}
\end{figure}

    \item \textbf{It is necessary to capture the spatio-temporal interdependence.}  NSTPP models the dependence of space on time by $p(s|t)$, and the performance regarding spatial metrics is improved compared with independent modeling, including Conditional KDE, CNF, and Time-varying CNF. However, it does not outperform other TPP models in the temporal domain, suggesting that modeling the distribution $p(t|H_t)$ without conditioning on the space $s$ fails to learn the temporal domain sufficiently. 

\begin{figure*}[t!]
    \centering
    \includegraphics[width=0.8\linewidth]{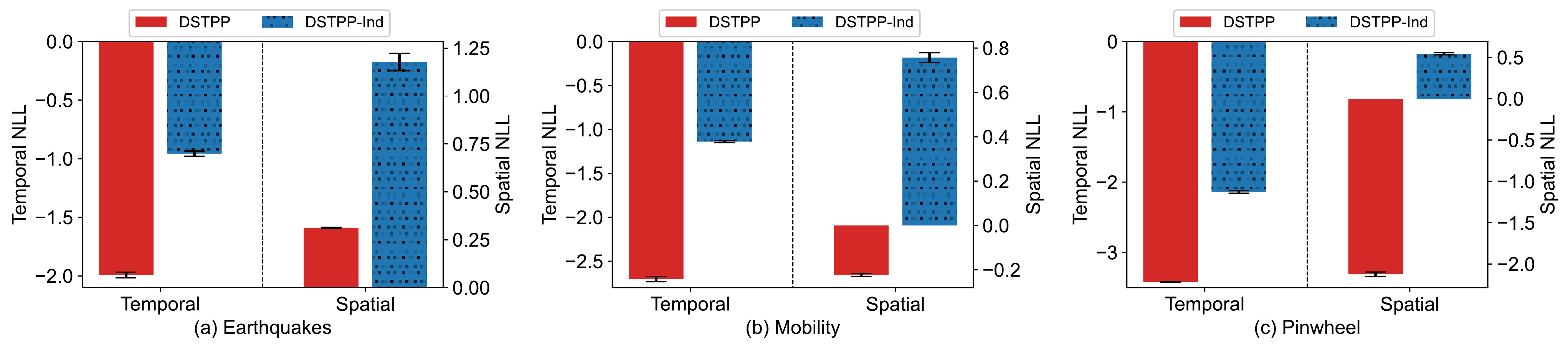}
    \vspace{-3mm}
    \caption{Ablation study on the joint spatio-temporal modeling. DSTPP-Ind denotes the degraded version of DSTPP, where spatial and temporal domains are independent. }
    \label{fig:joint}
\end{figure*}

    \item \textbf{DSTPP achieves the best performance across multiple datasets.} In the continuous-space scenarios regarding NLL, our model performs the best on both temporal and spatial domains. Compared with the second-best model, our model reduces the spatial NLL by over 20\% on average. The performance on temporal NLL also achieves remarkably significant improvement across various datasets.  
    In terms of models' predictive power, our model also achieves optimal performance, with remarkable improvements compared to the second-best model. In addition, as Figure~\ref{fig:discrete_main} shows, DSTPP delivers better predictive performance compared with other solutions in modeling discrete-space scenarios. The flexible framework that requires no parameter assumptions and MC estimations enables DSTPP to achieve superior performance.
\end{itemize}

\begin{figure}[t]
    \centering
    \includegraphics[width=0.95\linewidth]{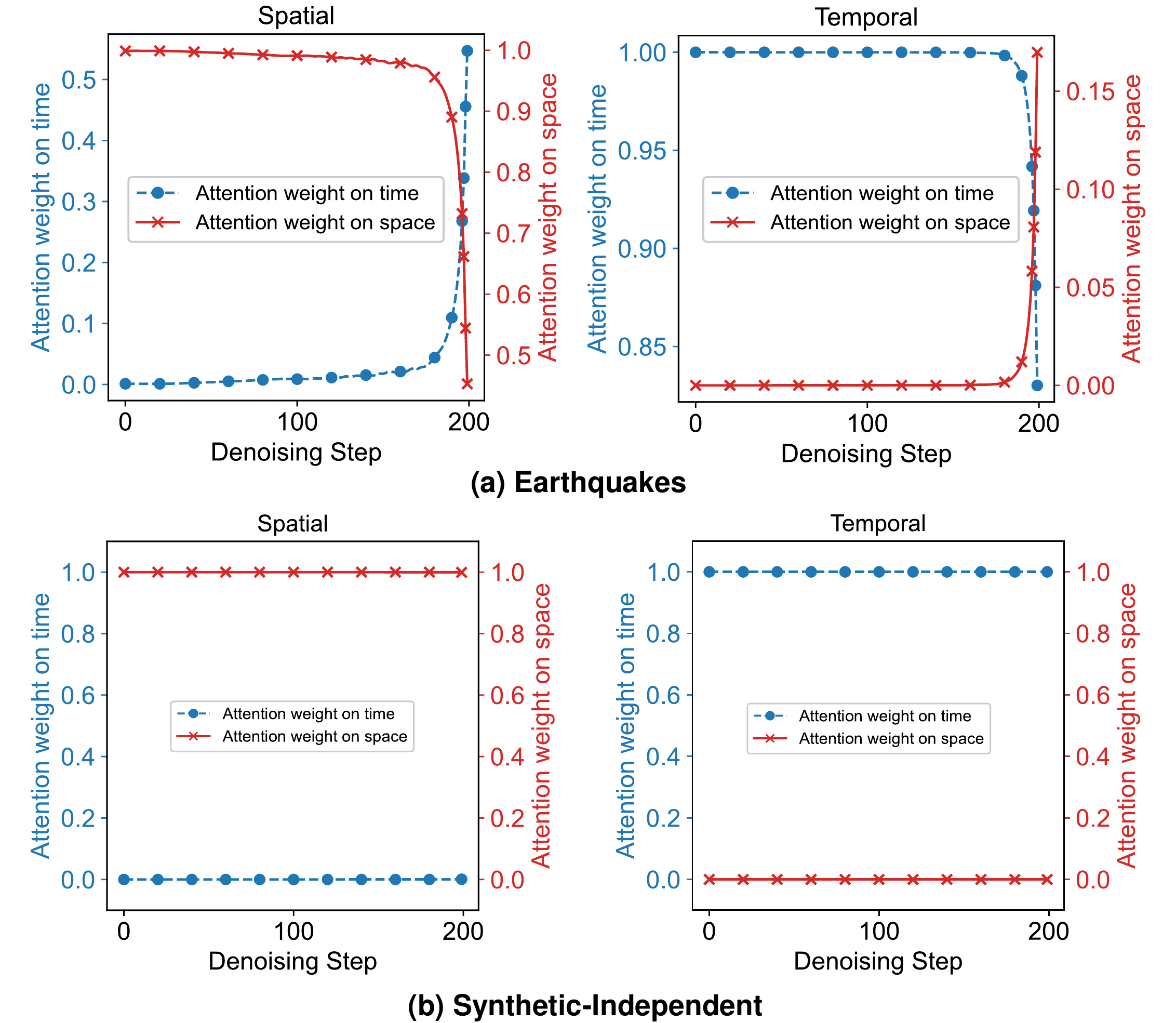}
    \caption{Spatial and temporal attention weights in the denoising iterations for two datasets with different spatio-temporal interdependence. Best viewed in color.}
    \label{fig:attn}
\end{figure}

\subsection{Analysis of Spatio-temporal Interdependence}\label{sec:attn}

To gain a deeper understanding of the spatio-temporal interdependence in the denoising process,  we perform an in-depth analysis of co-attention weights. Specifically, the analysis is conducted on two representative datasets: Earthquake and Synthetic-Independent, where the Earthquake dataset is highly spatio-temporal entangled, and the Synthetic-Independent dataset is totally spatio-temporal independent.  Appendix~\ref{sup:data} provides the generation details  of the synthetic dataset. 
We use these two datasets to validate whether the designed co-attention mechanism can learn different interdependence between time and space. 
At each step of the denoising process, we calculate attention weights of the temporal and spatial dimensions on  themselves and each other. Figure~\ref{fig:attn} shows how attention weights change as denoising proceeds.

As shown in Figure~\ref{fig:attn}(a), at the early stage, temporal and spatial domains do not assign attention weights to each other, and the attention weights on themselves are close to one. At the final stage (step $\geq 150$), the two domains start to assign attention weights to each other. At last, for the temporal domain, the attention weights on time and space are approximately 0.83 and 0.17;  for the spatial domain, the attention weights are close to evenly divided (0.52 and 0.48), suggesting that the spatial domain is more dependent on the temporal domain. 
In the later stage of the denoising iterations, the model learns a distribution closer to the real case; thus, it is reasonable that the spatial and temporal domains assign more attention weights to each other.
Figure~\ref{fig:attn}(b) displays different results: the two domains share almost no attention weights to each other, indicating that the model has successfully learned the independent relationship. Figure~\ref{fig:attn}(a) and (b) together validate the effectiveness of the co-attention mechanism, which can adaptively learn various interaction mechanisms between time and space.

\begin{figure*}[t!]
    \centering
    \includegraphics[width=0.88\linewidth]{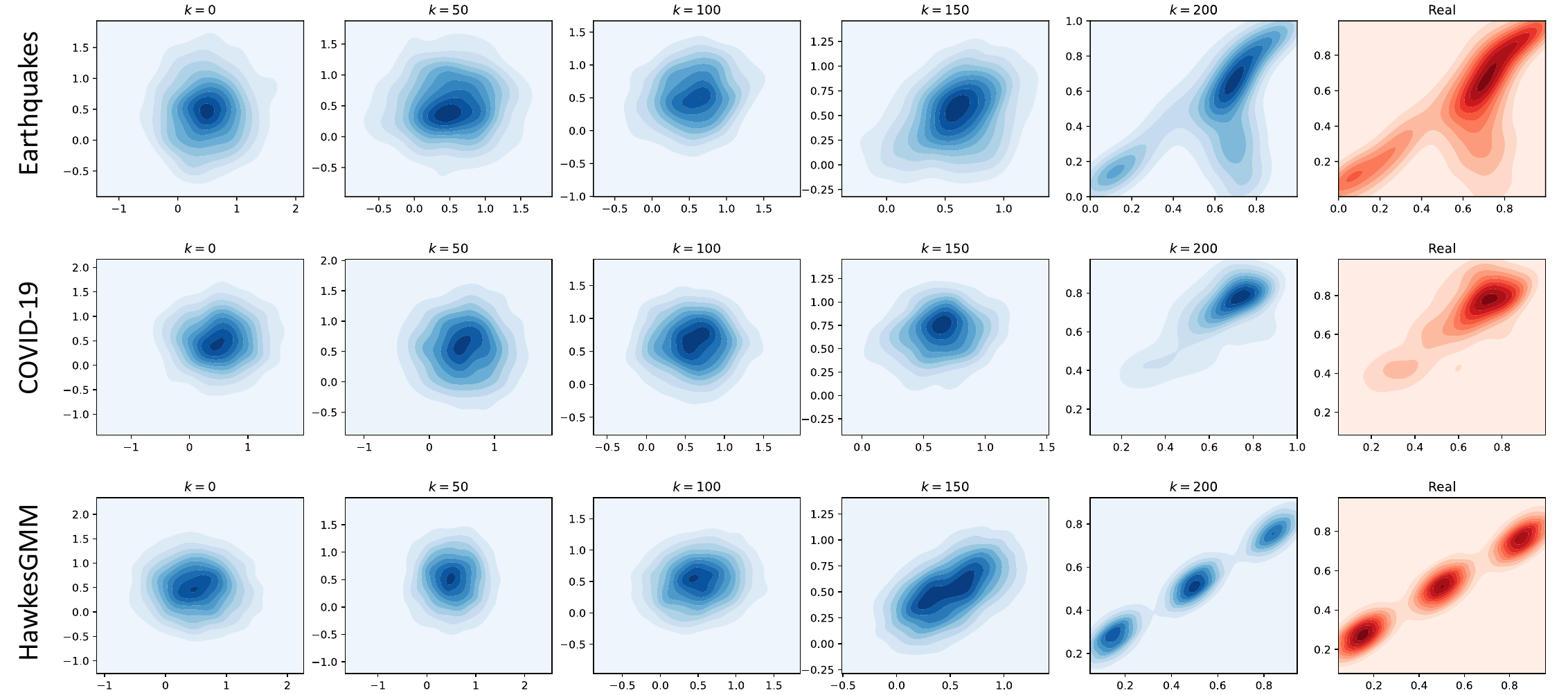}
    \caption{Visualization of the spatial distribution at different stages in the denoising process (the first five columns in blue color). The last column in red color presents the real distribution. Starting from Gaussian noise,  our DSTPP model gradually fits the spatial distribution of ground truth.  Best viewed in color. }
    \label{fig:denoise}
\end{figure*}

\subsection{Ablation Studies}

\textbf{Co-attention Mechanism.}
In order to examine the effectiveness of the co-attention mechanism, we degrade our DSTPP into a base framework, DSTPP-Ind, which models the distributions of space and time independently in the denoising process. To be specific, we replace $p_\theta(t_i^{k-1}|t_i^k, s_i^k, h_{i-1})$ and $p_\theta(s_i^{k-1}|t_i^k, s_i^k, h_{i-1})$ in Equation~\eqref{eq:st_joint} with $p_\theta(t_i^{k-1}|t_i^k, h_{i-1}), p_\theta(s_i^{k-1}|s_i^k, h_{i-1})$, where space and time are not conditionally dependent on each other.  Figure~\ref{fig:joint} shows the performance comparison of DSTPP and DSTPP-Ind in continuous-space settings. We can observe that DSTPP trained by incorporating the joint modeling of time and space performs consistently better than DSTPP-Ind with independent modeling. These results indicate the necessity to capture the interdependence between time and space, and meanwhile, validate the effectiveness of the spatio-temporal co-attention design.  Due to the space limit, we leave other results in Appendix~\ref{sup:add_result}.

\subsection{Analysis of Reverse Diffusion Processes}

To gain a deeper understanding of the denoising process, We visualize the spatial distribution during the reverse denoising iterations in Figure~\ref{fig:denoise}. 
As we can observe,  at the beginning of the denoising process, the spatial distribution displays a Gaussian noise. With progressive denoising iterations, the data distribution deforms gradually and becomes more concentrated. Finally, at the last step, the spatial distribution  fits perfectly with the ground truth distribution. It indicates that our DSTPP is able to learn the generative process of spatial distribution successfully. Besides, the denoising process is not a linear change, where the distribution changes during the last 50 steps are more significant than the previous steps. Combined with results in Section~\ref{sec:attn}, where the interdependence between spatial and temporal domains is effectively captured in the latter stage, it is reasonable that the denoising effect is improved significantly during this period.

\section{Related Work}

\textbf{Spatio-temporal Point Processes.}
Temporal point process models~\cite{du2016recurrent,mei2017neural,zuo2020transformer,zhang2020self,lin2022exploring} can be directly used for STPPs, where the space is considered as the event marker. Kernel density estimation methods are also used to model continuous-space distributions in  STPP models~\cite{chen2020neural, zhou2022neural,reinhart2018review,jia2019neural,moller2003statistical,baddeley2007spatial}.  Most existing solutions follow an intensity-based paradigm, and their main challenge is how to choose a good parametric form for the intensity function. There exists a trade-off between the modeling capability of the intensity function and the cost to compute the log-likelihood. Some intensity-free models~\cite{omi2019fully,shchur2019intensity} are proposed to tackle this problem; however, the probability density function either  is unavailable~\cite{omi2019fully} or still has certain model restrictions~\cite{shchur2019intensity}.  Another drawback of existing models is that they can only model either  the continuous-space domain or the discrete-space domain, which largely limits their usability in real-world scenarios.  

Recently, a line of advances have been developed for the generative modeling of point processes. For example, generative adversarial networks~\cite{xiao2017wasserstein, dizaji2022wasserstein} are used to learn to generate point processes in a likelihood-free manner. Reinforcement learning~\cite{li2018learning,upadhyay2018deep} approaches and  variational autoencoders~\cite{mehrasa2019variational,pan2020variational} are also included  to explore the generative performance of TPPs.  Some works also use noise contrastive learning~\cite{guo2018initiator,mei2020noise} instead of MLE. We are the first to learn point processes within the paradigm of diffusion models, which successfully address limitations in previous existing solutions.

\textbf{Denoising Diffusion Probabilistic Models.}
Denoising diffusion probabilistic models (DDPM)~\cite{sohl2015deep, ho2020denoising,song2020denoising}, are a class of deep generative models, which are inspired by non-equilibrium thermodynamics. Due to their powerful generative capabilities, diffusion models have been used in a wide range of applications including image generation~\cite{dhariwal2021diffusion,austin2021structured,sinha2021d2c,rombach2022high}, time series prediction and imputation~\cite{rasul2021autoregressive,tashiro2021csdi}, audio generation~\cite{goel2022s,ho2022video,kong2020diffwave}, text generation~\cite{li2022diffusion, gong2022diffuseq,gao2022difformer}, 3D point cloud generation~\cite{luo2021diffusion,zhou20213d}, and trajectory generation~\cite{gu2022stochastic, pekkanen2021variable}. In this paper, we first introduce the diffusion model to the domain of spatio-temporal point processes. 
\section{Conclusion}

In this paper,  we propose a novel framework to directly learn spatio-temporal joint distributions with no requirement for independence assumption and Monte Carlo sampling, which has addressed the structural shortcomings of existing solutions. The framework also poses desired properties like easy training and closed-form sampling.  Extensive experiments on diverse datasets highlight the impact of our framework against state-of-the-art STPP models. 
As for future work, it is promising to apply our model in urban system~\cite{yu2022spatio, li2022automated} as well as
large-scale natural systems, such as climate changes and ocean currents, which are concerned with highly complex spatio-temporal data.

\begin{acks}
 This work was supported in part by the National Key Research and Development Program of China under grant 2020YFA0711403, the National Nature Science Foundation of China under U22B2057, 61971267, and U1936217, and BNRist. 
\end{acks}

\clearpage

\bibliographystyle{ACM-Reference-Format}
\bibliography{reference}


\section*{Appendix}

\appendix

\section{Dataset}\label{sup:data}

\textbf{Earthquakes:} Earthquakes in Japan from 1990 to 2020 with a magnitude of at least 2.5 are collected from the U.S. Geological Survey\footnote{https://earthquake.usgs.gov/earthquakes/search/}. Sequences are generated by sliding windows with a window size of 30 days and a gap of seven days.  Therefore, each sequence is within the length of  30 days. The earthquakes from Nov. 2010 to Dec. 2011 are removed because they are outliers compared with data in other periods. We split the dataset into the training set, validation set, and testing set and ensure that there is no overlap between them.  Finally, We have 950 sequences for the training set, 50 for the validation set, and 50 for the testing set. The sequence lengths range from 22 to 554. 

\textbf{COVID-19:}  We construct this dataset by using publicly released COVID19 cases by The New York Times (2020). It records daily infected cases of COVID-19 in New Jersey state\footnote{https://github.com/nytimes/covid-19-data} from March 2020 to July 2020. We aggregate the data at the county level. Sequences are generated by sliding windows with a window size of 7 days and a gap of three days. Therefore, each sequence is within the length of  7 days. We split the dataset into the training set, validation set, and testing set and ensure that there is no overlap between them.  Finally, We have 1450 sequences for the training set, 100 for the validation set, and 100 for the testing set. The sequence lengths range between 5 to 287.


\textbf{Citibike:} This dataset is collected by a bike-sharing service, which records the demand for bike sharing in New York City. We use the records from April 2019 to August 2019. The start of each trip is considered as an event. We split the record sequence of each bike into one-day subsequences starting at 5:00 am in the day. Therefore, each sequence is within the length of  19 hours. We randomly split the dataset into the training set, validation set, and testing set. Finally, We have 2440 sequences for the training set, 300 for the validation set, and 320 for the testing set. The sequence lengths range from 14 to 204.

\textbf{Crime~\footnote{http://www.atlantapd.org/i-want-to/crime-data-downloads}:} It is provided by the Atlanta Police Department, recording robbery crime events from the end of 2015 to 2017. Each robbery report  is associated with the time and the neighborhood. Each sequence is within the length of  one day. We randomly split the dataset into the training set, validation set, and testing set. Finally, We have 2000 sequences for the training set, 200 for the validation set, and 2000 for the testing set. The sequence lengths range between 26 to 144.

\textbf{Synthetic-Independent:} The temporal domain is generated by a Hawkes process, and  the intensity function is defined as follows:

\begin{equation}
    \lambda(t,|H_t)=0.2+\sum_{t<t_i}(0.2e^{-0.2(t-t-t_i)}+4e^{-10(t_i-t)})
\end{equation}

\noindent the spatial distribution follows a Two-dimensional Gaussian distribution:

\begin{align}
    & f(s1,s2)= \\
    & \frac{1}{2\pi\sigma_{s1}\sigma_{s2}\sqrt{(1-\rho^2)}}e^{-\frac{1}{2(1-\rho^2)}[(\frac{s1-\mu_1}{\sigma_{s1}})^2-2\rho(\frac{s1-\mu_1}{\sigma_{s1}})(\frac{s2-\mu_2}{\sigma_{s2}})+(\frac{s2-\mu_2}{\sigma_{s2}})^2]}
\end{align}

\noindent where $\rho = \frac{\sqrt{2}}{4}, \mu_1 = 4.0, \mu_2 = 7.0, \sigma_{s1} = \sqrt{2}, \sigma_{s2} = 2$.

\section{Baseline}\label{sup:baseline}
We provide detailed descriptions of used baselines as follows: 

\begin{itemize}[leftmargin=*]
    \item Conditional KDE: Conditional kernel density estimations. We utilize a history-dependent Gaussian mixture model to model the spatial distribution. 
    \item CNF and Time-varying CNF~\cite{chen2018neural}: We use Continuous normalizing flow for modeling spatial distribution.  Time-varying CNF denotes the dependence on the timestamps. 
    \item Possion~\cite{rasmussen2018lecture}: The homogeneous Poisson process is the simplest point process, where the number of events occurring during time intervals are independent, and the probability of a single event occurrence is proportional to the length of the interval. 
    \item Hawkes~\cite{hawkes2018hawkes}: Its essential property is that the occurrence of any event increases the probability of further events occurring by a certain amount. The triggering kernel which captures temporal dependencies can be chosen in advance or directly learned from data. 
    \item Self-correcting~\cite{isham1979self}: In contrast to the Hawkes process, this point process follows the pattern that the occurrence of past events inhibits the occurrence of future events. Every time when a new event appears, the intensity is decreased by multiplying a constant less than 1.
    \item Recurrent Marked Temporal Point Process (RMTPP)~\cite{du2016recurrent}: This neural temporal point process model applies a nonlinear function of past events to model the intensity function and leverages RNNs to learn a representation of the influence from event history, where time intervals act as explicit inputs. 
    \item Neural Hawkes Process (NHP)~\cite{mei2017neural}: With the goal of capturing the temporal evolution of event sequences, it uses continuous-time LSTMs to model a marked TPP. The modeling of future event intensities is conditioned on the RNN's hidden state.
    \item Transformer Hawkes Process (THP)~\cite{zuo2020transformer}: It is an extension to the transformer by modeling the conditional intensity. The self-attention mechanism is leveraged to capture long-term dependencies.
    \item Self-attentive Hawkes Process (SAHP)~\cite{zhang2020self}: It learns the temporal dynamics by leveraging a self-attention mechanism to aggregate historical events. In order to take the time intervals between events into consideration, it modifies the conventional positional encoding by converting time intervals into phase shifts of sinusoidal functions.
    \item Log Normal Mixture model (LogNormMix)~\cite{shchur2019intensity}: It adopts intensity-free learning of TPPs, which models the PDF by a log-normal mixture model. Additionally, a simple mixture model is proposed to match the flexibility of flow-based models. The loglikelihood for training and density for sampling are in closed form.
    \item Wasserstein GAN (WGAN)~\cite{xiao2017wasserstein}: This intensity-free approach transforms nuisance processes to target one. And the Wasserstein distance is used to train the model, which is a likelihood-free method. Loglikelihood cannot be obtained for this approach. 
    \item Neural Jump Stochastic Differential Equations (NJSDE)~\cite{shchur2019intensity}: It models TPPs with a piecewise-continuous latent representation, where the discontinuities are brought by stochastic events. The spatial distribution is modeled with a Gaussian mixture model. 
    \item Neural Spatio-temporal Point Process (NSTPP)~\cite{chen2020neural}: It applies Neural ODEs as the backbone, which parameterizes the temporal intensity with Neural Jump SDEs and the spatial PDF with continuous-time normalizing flows. 
    \item Deep Spatiotemporal Point Process (DeepSTPP)~\cite{zhou2022neural}: It is the state-of-the-art STPP model, which suggests using a non-parametric space-time intensity function governed by a latent process. Amortized variational inference is leveraged to deduce the latent process. 
\end{itemize}

\section{Implementation Details}\label{append:implementation}

\subsection{Evaluation Metrics}\label{sup:metric}

Suppose $\bm{y}=y_1, ..., y_M$ represents the ground truth for real values, $\bm{\hat{y}}=\hat{y}_1, ..., \hat{y}_N$ represents the predicted real values, $\bm{k}=k_1, ..., k_M$ represents the ground truth for real values, $\bm{\hat{k}}=\hat{k}_1, ..., \hat{k}_N$ represents the predicted discrete labels, and $N$ denotes the number of test samples, we can formulate these metrics as follows:

\begin{equation}
\begin{split}
    &\mathrm{RMSE}(\bm{y}, \bm{\hat{y}}) = \sqrt{\frac{1}{N}\sum_{i}^{N} \left ( y_i - \hat{y}_i \right )^2}, \\
    &\mathrm{d_{Euclid}}(\bm{y}, \bm{\hat{y}}) = \frac{1}{N}\sum_{i=1}^N\|y_i-\hat{y}_i\|\\
    & \mathrm{Accuracy}(\bm{k}, \bm{\hat{k}}) = \frac{TP+TN}{TP+TN+FP+FN}
\end{split}
\end{equation}

\noindent where $\mathrm{d_{Euclid}}$ denotes the Euclidean distance between two real-valued vectors. TN, TP, FN, and FP are the number of true negative samples, the number of true positive samples, the number of false negative samples, and the number of false positive samples. 

\subsection{Parameter Settings}\label{sup:param}

We all use three-layer MLPs with ReLU activations and hidden size of 64. The training process is performed in batch. After training the model for ten epochs (~all training instance), we examine the model's performance on validation set. The model that delivers the best performance on the validation set will be used to validate the performance on the test set. We set the learning rate as 3e-4 via searching in a set of $\{1e-4,3e-4,1e-3\}$. 
The proposed framework is implemented with Pytorch. We train it  on a Linux server with eight GPUs (NVIDIA RTX 2080 Ti * 8). In practice, our framework can be effectively trained within 6 hours on a single GPU. 

\begin{figure}[t]
    \centering
    \includegraphics[width=0.95\linewidth]{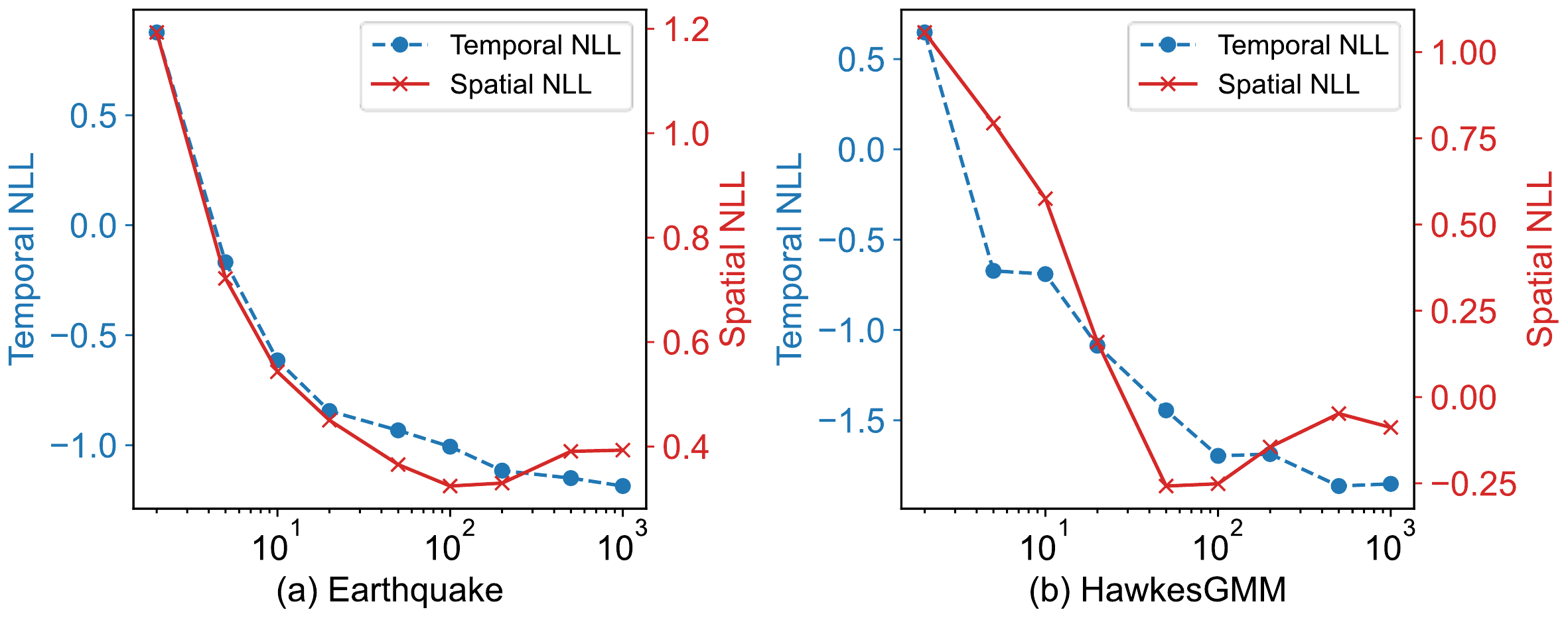}
    \caption{Ablation studies on the total number of diffusion steps for Earthquake and HawkesGMM data. We observe similar results with other datasets.}
    \label{fig:ablation_T}
    \vspace{-3mm}
\end{figure}

\section{Additional Results}\label{sup:add_result}

\textbf{Diffusion Steps.}
The number of total steps $K$ in the diffusion process is a crucial hyperparameter. With the increase of diffusion steps, the denoising network approximates more minimal changes between steps. A bigger $K$ allows the reverse denoising process to be adequately approximated by Gaussian distribution~\cite{sohl2015deep}. However, too many diffusion steps will vastly reduce the efficiency of training and sampling. Therefore, it is essential to explore  to what extent, a larger diffusion step $K$ improves the model's performance.  Specifically, we perform ablation studies on Earthquakes and HawkesGMM datasets with varying total diffusion steps $K = \{2,5,10,20,50,100,200,500,1000\}$ and keep all other hyperparameters fixed. The results of temporal NLL and spatial NLL are plotted in Figure~\ref{fig:ablation_T}. We can observe that temporal NLL and spatial NLL both achieve the best values at $K\approx 200$, suggesting that the diffusion step $K$ can be reduced to 200 without significant performance loss.




\end{document}